# Fast and Accurate Identification of Hardware Trojan Locations in Gate-Level Netlist using Nearest Neighbour Approach integrated with Machine Learning Technique

Anindita Chattopadhyay [a], Siddharth Bisariya [b], Vijay Kumar Sutrakar [c, *]

[a]*Dept. Of Electronics and communication, BMS College of Engineering, Bangalore, India*

[b]*Department of Electrical and Electronics, SCOPE College of Engineering, Bhopal, Madhya Pradesh, India*

[c]*Aeronautical Development Establishment, Defence Research and Development Organisation, Bangalore, India*

** Corresponding Author Email: vks.ade@gov.in*

**Abstract**

In the evolving landscape of integrated circuit (IC) design, detecting exact locations of Hardware Trojans (HTs) within a multi-entity-based design cycle present significant challenges. Also, accurate detection of exact locations of HTs in a complex large IC design will lead to significant amount of time saving. The present research work proposes an innovative idea of using machine learning-based methodology for identifying malicious logic gates in gate-level netlists by focusing on path retrace algorithms. Initially, three different machine learning methodologies are employed for the identification of HTs in the circuits. Case-I utilises a decision tree algorithm for node-to-node comparisons, significantly improving detection accuracy through the integration of Principal Component Analysis (PCA). Case-II introduces a graph-to-graph classification using a Graph Neural Network (GNN) model, enabling the differentiation between normal and Trojan-infected circuit designs with an accuracy of 62.8%. Case-III applies GNN-based node classification to identify individual compromised nodes and their location with an accuracy of 79.8%. In order to further improve the identification of HTs and model accuracy, nearest neighbour (NN) approach has been combined with the GNN graph-to-graph (Case II) and GNN node-to-node (Case III). Integration of GNN model graph-to-graph classification leads to improvement in accuracy from 62.8% to 73.2% and 97.7% with 1$^{st}$ NN and 2$^{nd}$ NN approaches, respectively. It leads to minimum time saving of ~93% and ~50% for HT detections in Case II using 1$^{st}$ NN and 2$^{nd}$ NN models, respectively. Similarly, GNN model node-to-node classification leads to improvement in accuracy from 79.8% to 93% and 97.7% with 1$^{st}$ NN and 2$^{nd}$ NN approaches, respectively. GNN model node-to-node classification leads to minimum time saving of ~92% and ~50% for HT detections in Case III using 1$^{st}$ NN and 2$^{nd}$ NN models, respectively. Result shows significant improvement in accuracy and fast detection of HTs using NN approach integrated with GNN models.

## 1. Introduction

Globalization has brought significant transformations in the entire supply chain of integrated circuits (ICs). In the contemporary landscape of computer systems, chips may undergo manufacturing processes including fabrication and assembly in various locations around the world. This imposes the challenge of tracing the origin of these components [1]. Moreover, the heavy reliance on overseas foundries for IC fabrication, coupled with the growing awareness of the vulnerability of chips to HT insertion in these foundries, has intensified security concerns. In the recent past, several methods have been proposed for HT detections for ensuring the trustworthiness of the fabricated chips [2-4]. However, the primary obstacle encountered in previously developed non-destructive approaches is the accuracy of detection.

Consequently, to tackle the challenge of detecting HTs, it becomes necessary to perform chip-level reverse engineering. However, this process is very expensive. Nowadays, advanced imaging-based techniques are a promising option for driving the complete netlist from a fabricated chip [5]. This also helps the designers for comparing the changes with their original design and identifying any changes, if any, incorporated during the chip fabrication process. However, such techniques are not applicable to Commercial Off-The-Shelf (COTS) ICs and System-on-Chips (SoCs) incorporating third-party Intellectual Properties (IPs), due to the absence of golden models. COTS ICs and SoCs require more sophisticated design analysis tools. Recent developments in these area have seen the emergence of various tools [6]. However, these tools consider all the internal registers uniformly and attempt to recognize functional logic through pattern matching, resulting in a mixture of control logic and data path within the identified functional components. This complicates the task for testers in fully recovering the design's functionality. Hence, identification and pin-pointing the exact locations of functional Trojans inside ICs are still a major area of concern.

In this paper, an attempt has been made to identify functional Trojans (especially their exact locations inside the circuits) that are inserted maliciously during the fabrication process. The proposed approach demonstrates a comprehensive methodology for detecting HTs in digital circuits, leveraging graph-based representation and machine learning techniques as shown in figure 1. The process begins with a gate-level netlist, which contains the digital circuit in terms of logic gates and their interconnections. This netlist is then flattened, simplifying the design into a single level of logic gates and connections. The flattened netlist is subsequently transformed into a graph representation. This graph representation facilitates the application of GNNs and other machine learning models to detect anomalies indicative of HTs. The output from the proposed model identifies nodes and subgraphs that are potentially infected with HTs. Finally, the process involves locating the exact positions of these infected nodes within the original gate-level netlist, enabling precise identification of compromised parts of the circuit and facilitating subsequent corrective actions. The designed netlist of logic circuits serves as the reference. A path retrace algorithm is employed to pin-point the locations of malicious nets within the circuit [6]. This integrated approach combines traditional circuit design with advanced machine learning, leading to (a) enhancing the detection and mitigation of HTs and (b) ensuring the safety and truthfulness of digital systems. Further attempts have also been made for improving the accuracy of the machine learning models utilizing nearest neighbour (NN) approach.

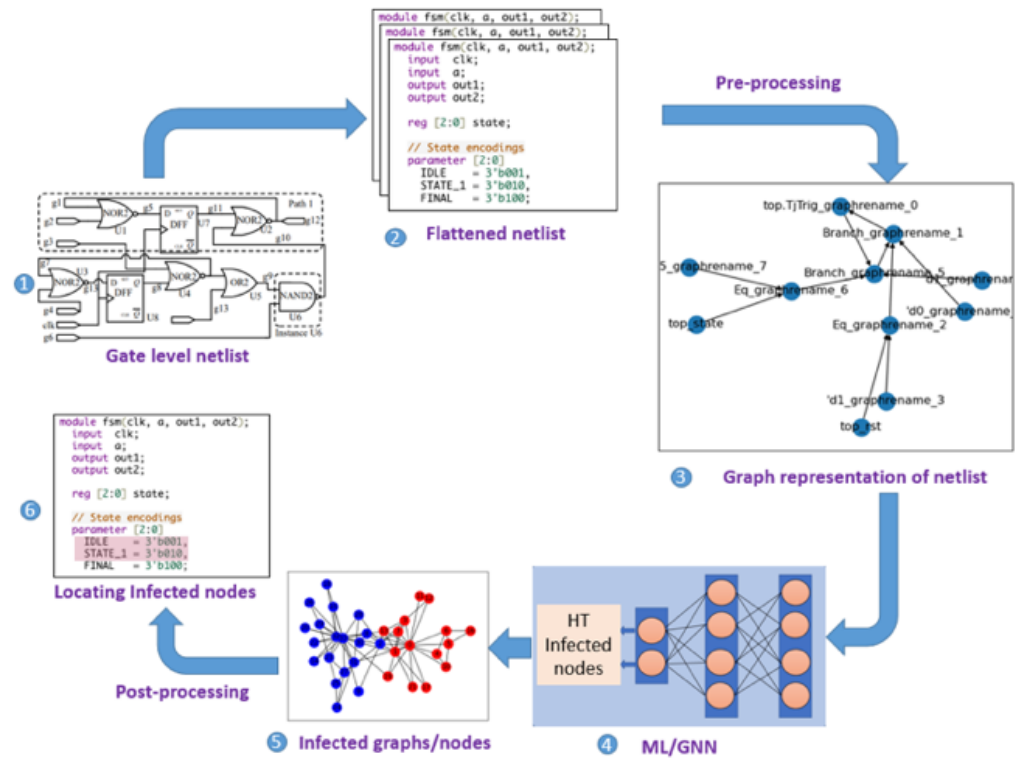

**Fig. 1.** ML method employed for HT detection and identification of its location inside the netlist

The subsequent sections of the paper are organized as follows: Section II provides an overview of current work related to state register identification and netlist classification. Section III offers a brief overview of the proposed methodology. In section IV, data sets and the details of the proposed model are presented. Finally, Section V concludes the paper with a summary and suggests potential applications of the proposed methodology.

## 2. Related Work

The design of digital circuits commonly occurs at the Register Transfer Level (RTL). RTL provides the details of signal flow among various registers. RTL descriptions are then transformed into gate-level netlists utilizing logic synthesis tools, outlining gates and their interconnections. Subsequently, place and route algorithms are used for the placement of gates and routing of interconnections. However, this translation incurs information loss from a reverse engineering standpoint (from netlist to RTL), particularly in terms of module boundary and hierarchy information [7][8]. Moreover, various optimizations are implemented to achieve predefined goals like improved area or timing. Consequently, reverse engineers target their data to recover essential high-level design information [9-14].

A comprehensive case study is presented to provide an exhaustive review of the intersection between machine learning (ML) techniques and hardware security, specifically focusing on the detection and mitigation of hardware Trojan attacks using supervised learning techniques have been prominently discussed for their high accuracy in classifying known Trojan signatures [9]. A novel technique that enhances the efficiency of extracting FSMs is also proposed in the past that is very helpful in understanding the behaviour of digital circuits [10]. Building upon previous work, the research work delves into the extraction of functional modules from flattened gate-level netlists. The research expands understanding of identifying and categorizing functional components within digital circuits, offering valuable insights for reverse engineering practitioners [11]. The behavioural pattern mining-based reverse engineering circuits technique is also explored in the past. It provides the application of pattern mining in reconstructing circuit behaviors, contributing to the advancement of effective reverse engineering methodologies [12] [13]. A template-based solution is also proposed to the challenges posed by circuit complexity, contributing to the arsenal of tools available for reverse engineers [14]. Previous approaches encounter difficulties in fully recovering module boundaries and hierarchy from gate-level netlists due to information loss during synthesis. To address this, the proposed work introduces an innovative method for more accurate and exact locations of HT inside complex ICs.

## 3. Methodology

In the recent past, machine learning techniques have been used extensively for HT detections. Among these, GNNs have shown promising results due to their ability to effectively model the complex interconnections within a circuit represented as a netlist [15]. By treating the circuit as a graph, where nodes represent individual components (gates) and edges (signals) represent their connections, GNNs can capture intricate relationships and patterns indicative of Trojan behaviour.

In general, HT detection in netlists using GNNs uses the following approach: At first, the netlist data is pre-processed to extract relevant features and construct the graph representation. Next, designed a GNN architecture tailored to capture the unique characteristics of netlist graphs while mitigating issues such as overfitting and computational complexity. Subsequently, the model is trained using labelled datasets containing both Trojan-free and Trojan-infected netlists. Finally, the performance is evaluated of the trained model. Matrices such as accuracy, precision, recall, and F1-score are generally used for assessing the efficiency of machine learning models in Trojan detections. The GNN methodology holds the potential to complement existing detection techniques and pave the way for more robust and scalable solutions in safeguarding electronic systems against hardware-level threats.

In this research work, a technique based on the GNN approach is proposed. It is based on a machine learning framework designed to autonomously and precisely identify and categorize Trojans inside the circuits within a flattened netlist. The core concept revolves around utilizing GNNs capabilities for understanding and analyzing both the structural and functional attributes of different circuits. This initiative stems from the recognition that a gate-level netlist can be depicted as a graph, where individual sub-circuits or sub-graphs typically display unique functionalities and configurations. Machine learning driven HT detection techniques categorize Trojan circuits based on features. To gather feature extraction, established HT benchmarks are obtained from Trust-Hub [16]. The netlist is generated using Yosys [17].

The implementation of the proposed strategy involves several key steps. Initially, the gate-level netlist data, representing the structural elements of the integrated circuit design, is collected and pre-processed to extract relevant features. These features may include connectivity patterns, gate types, and other attributes indicative of potential Trojan activity. Subsequently, a ML algorithm, such as a decision tree or neural network, is trained on labelled data to learn patterns associated with both normal circuits and circuits with Trojan insertions. During the training phase, emphasis is placed on incorporating techniques to handle class imbalance and optimize model performance. Once the model is trained and validated, it is deployed to analyze new gate-level netlists, where it identifies suspicious patterns indicative of Trojan presence. The proposed strategy enables automated and efficient detection of HTs and its locations inside the gate-level netlist, offering a proactive defense mechanism against malicious modifications in integrated circuit designs. Through continuous refinement and validation, this approach holds promise for bolstering the security of hardware systems against emerging threats in the semiconductor industry.

## 4. Dataset and Model

The proposed model employs a set of benchmarks comprising gate-level netlists taken from Trust-Hub [16]. In order to make datasets more generalized, additional netlists of new data are also generated using Yosys [17]. These netlists contain a variety of HTs, including combinational and sequential logic [15-19]. The devised HT feature extraction approach is implemented using Python. This model is then employed for training and testing with the extracted features derived from HT benchmarks sourced from (a) Trust-Hub [16] and (b) the netlist generation platform in Yosys [17]. The dataset used for hardware Trojan detection is sourced primarily from the Trust-Hub platform, specifically from the chip and netlist sections. It includes a variety of circuit designs that encompass a range of complexities and potential vulnerabilities. The dataset is composed of 1025 circuits, with a mixture of both standard designs and those intentionally embedded with hardware Trojans. The Trust-Hub Database serves as a widely used benchmark for HT detection, featuring diverse circuits, including AES, RS232, memctrl, PIC16F84, WB_CONMAX, and a simple RISC-V processor. These circuits range from small controllers to larger system-on-chip (SoC) designs, covering both combinational and sequential logic. Each circuit has multiple versions (clean and Trojan-infected) ensuring a comprehensive evaluation framework. The dataset includes various Trojan types, categorized as functional (logic-altering), parametric (power or delay-based), and structural (modifying connections or gates). AES contains multiple Trojan variants, while RS232 incorporates nine distinct Trojans. These Trojans employ three triggering mechanisms: input data-triggered HTs (activated by specific input patterns), normally open Trojans (active throughout operation), and state-based HTs (triggered by predefined internal states, such as a counter value). Beyond Trust-Hub, the RISC-V, Web3 Hardware and custom created hardware description language (HDL) HT dataset generated using Yosys (as a part of present work) expands the benchmark with 290 Verilog-based designs, including ASIC-derived feature-extracted data for ML-based classification.

The proposed methodology provides a solid foundation for developing and evaluating machine learning models, particularly in detecting HTs across different circuit architectures. An extensive evaluation process encompassing three distinct cases to assess the efficacy of this proposed methodology is employed. In Case I, a machine learning model is employed for evaluating the performance of the proposed approach under standard conditions. This case serves as a reference benchmark for comparison with existing techniques that operate solely at the node level. In Case II, graph classification is considered to evaluate the accuracy and efficiency of the proposed model. Case III, utilize a GNN model for node classification. This comprehensive evaluation strategy allows for a thorough examination of the capabilities of proposed methodology across different scenarios, ensuring a comprehensive assessment of its performance and effectiveness.

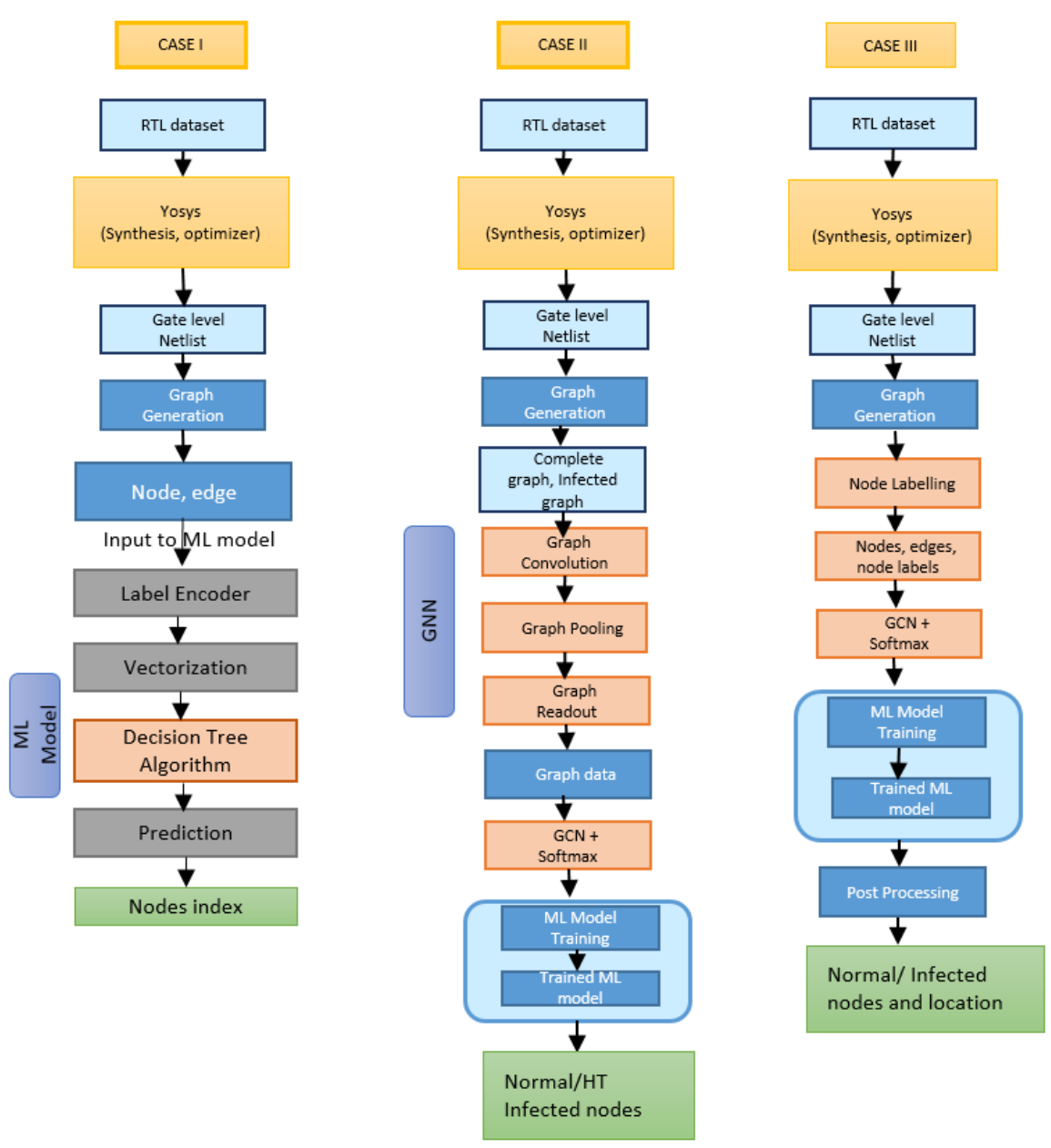

**Fig. 2.** The proposed Methodology

*Case I: HT detection using machine learning*

The proposed model in case I presents a structured workflow for detecting HTs in digital circuits using an ML technique, specifically focusing on a decision tree algorithm (refer Figure 2 for further details). In Yosys, the RTL description undergoes synthesis and optimization to generate a gate-level netlist. The process begins with frontend parsing, where Yosys reads the Verilog or VHDL code and converts it into an intermediate representation (IR). Next, logic synthesis is performed, mapping high-level constructs into primitive gates such as AND, OR, and flip-flops. The technology mapping stage then optimizes the design by selecting appropriate standard cells based on the synthesis library. Finally, the flattening and netlist export stage produces a structural gate-level netlist, which is then used for graph-based Trojan detection, where gates become nodes and connections form edges in the extracted circuit graph.

The next step involves preparing the data for machine learning. Each node and edge in the graph are encoded with labels to facilitate further analysis. The encoded data undergoes vectorization and transforming it into a numerical format suitable for machine learning models. The core of the process employs a decision tree algorithm. It is used for analysing the vectorized data and recognize patterns or anomalies suggestive of HTs.

The objective of the proposed technique is to improve the accuracy and efficiency of Trojan detection by analysing the internode relationships within the circuit netlist. By iteratively comparing nodes and leveraging decision tree-based classification, this approach facilitates the identification of potential HTs within the circuitry. This methodology integrates traditional electronic design automation (EDA) tools with advanced machine learning techniques to enhance the detection of malicious hardware modifications, ensuring the security and integrity of digital systems. By leveraging the decision tree algorithm, the approach provides a robust framework for identifying HTs, contributing to the field of hardware security.

**Algorithm 1:** Decision Tree for Trojan detection in netlist

**Input:** Gate level Netlist files (Verilog) of circuits *.v*
**Output:** Decision ($D$) = Trojan Infected node or Trojan Free node
for each circuit *.v* do;
    $g$ ← HW2GRAPH (*.v*);
    $h_g$ ← GRAPH2VEC ($g$);
    $\hat{y}$ ← Decision Tree ($h_g$);
    If $\hat{y}$ == 0 then
        return NON_TROJAN;
    else
        return TROJAN

In this case, a decision tree algorithm is employed for node-to-node comparisons. This method is tested using PCA and t-SNE for evaluating the improvement in accuracy. The approach of integrating PCA significantly improves detection accuracy compared to t-SNE in the present case. PCA is chosen for dimensionality reduction due to its computational efficiency and ability to retain key variance in high-dimensional feature spaces. Before applying PCA, feature vectors had 256 dimensions, which are reduced to 50 principal components to balance computational efficiency and information retention. This selection ensures minimal loss of critical features while improving processing speed.

The decision tree method effectively reduces the dimensionality of the feature space, allowing for more accurate comparisons between nodes. However, this approach has limitations. For instance, it may not be effective in detecting complex Trojan designs or identifying the exact locations of the compromised nodes at the individual level. Decision trees are suitable for simple, binary classification problems. However, it may struggle with more nuanced, multiclass scenarios. Therefore, using GNNs provides a more robust approach, capturing complex relationships and improving detection accuracy in intricate Trojan designs.

*Case II: graph to graph comparison*

To address the limitations of Case I, a graph-to-graph classification approach using a Graph Neural Network (GNN) model is introduced as Case II. This method enables differentiation between normal and Trojan-infected circuit designs by leveraging the structural properties of graphs. The GNN model can effectively capture complex patterns and relationships within circuit designs, making it well-suited for detecting sophisticated Trojans.

Case II illustrates a workflow for generating graph data from RTL datasets using the Neural GNN model (refer Figure 2 for further details). Initially, RTL datasets are synthesized and optimized using Yosys, producing a gate-level netlist. This netlist undergoes graph generation to form a complete graph, including infected graph data, which represents potential errors or modifications. The methodology involves transferring a given netlist into graph representation where nodes correspond to logic gates and edges define them the interconnects. A database of identified infected graph containing Trojan patterns is created for comparative analysis.

To identify Trojan location, the netlist graph is systematically compared against infected graphs using graph intersection technique. When a subgraph within the netlist exhibits a significant overlap with a non-infected graph, it is flagged as potential Trojan affected region. The nodes involved in this intersection are extended and mapped back to the original netlist for pin point the exact location.

The GNN model then processes the graph through three main stages, i.e., (a) graph convolution, (b) graph pooling, and (c) graph readout. Graph convolution extracts feature from the graph, graph pooling reduces the graph's complexity while preserving essential information, and graph readout generates the final graph data output. These labelled graphs are passed through a Graph Convolutional Network (GCN) with a Softmax activation function to extract features and perform node classification. The foundation of methodology lies in formulating the HT detection problem as a binary classification task, where each circuit instance is classified as either benign or Trojan-infected. A comprehensive dataset comprising both Trojan free and Trojan-infected circuits is then constructed, ensuring a diverse and representative training corpus.

The resulting graph data, comprising nodes, edges, and their labels, is then utilized to train the machine learning model. This trained ML model is subsequently used to identify HT-infected nodes, yielding detection results that differentiate between normal and HT-infected nodes. This approach leverages the graph's structural information and the ML models' learning capabilities to efficiently detect and classify potential HTs within the dataset. Thus, the model demonstrates the efficacy of the deep learning-based approach in accurately detecting HTs. Hence, surpassing traditional methods in terms of both detection accuracy and computational efficiency, thus safeguarding electronic systems against potential threats posed by malicious hardware modifications. Implementation details are provided in Algorithm 2.

The model also incorporates nearest neighbour (NN) techniques, examining both the 1st and the 2nd NNs during the classification process. By taking into account each node's immediate neighbour, i.e., 1st nearest and the next closest to the 1st nearest neighbour, i.e., 2nd nearest, the model enhances its capability to detect HTs. This layered neighbour analysis provides additional context, enabling more accurate identification of Trojan related irregularities. By leveraging structural similarity between infected and clean circuits the approach reduces the dependency on heuristic-based detection which can be prone to high false positive rate. Furthermore, the proposed approach is even more effective for large scale circuits where exhaustive search technique is computationally impractical.

**Algorithm 2:** Trojan Detection with Nearest Neighbour Analysis

Input: Gate-level Netlist files (Verilog) of circuits (.v)
Output: Trojan Infected Node or Trojan-Free Node with Nearest Neighbour Analysis

```
def Trojan_location_detection(netlist_files):
    for each circuit.v in netlist_files do:
        g ← HW2GRAPH (.v)
        hg ← GRAPH2VEC (g)
        ŷ ← GCN (hg)
            # Generate infected graph representation
              infected_graphs = generate_infected_graphs()
            # Graph intersection to detect Trojan affected subgraph
              intersection_graph = graph_intersection(g, infected
graph)
            if intersection_graph:
             #identify specific nodes involved in intersection
               Trojan_nodes = extract_trojan_nodes(intersection_graph)
             # Map detected nodes to original netlist
               Trojan_locations = map_to_netlist (Trojan_nodes, circuit)
             return f " TROJAN detected in nodes:{ Trojan_location}"
return "NON_TROJAN"

        for node in g.nodes:
            NN1 ← find_nearest_neighbour (node, level=1)
            NN2 ← find_nearest_neighbour (NN1, level=2)
            f_i ← extract_features (node, NN1, NN2)
            y_node = GCN (f_i)
            if y_node == 'TROJAN':
                 return f"TROJAN detected in node: {node}"

   return "NON_TROJAN"
```

While Case II enhances the capabilities compared to Case I, there are still scope for improvements, particularly in identifying individual compromised node locations within a circuit design.

*Case III: GNN model using node classification*

Building on the strengths of Case I and Case II, Case III enhances detection by applying GNN-based node classification, enabling precise identification of individual compromised nodes and their locations. This approach leverages the graph structure of a circuit design to predict node labels (normal or infected). By applying this method at the node level, the extent and location of compromise within a circuit design can be gained.

Case III presents a detailed workflow for detecting HT in RTL datasets using a combination of graph-based methods and machine learning techniques. The process begins with the RTL dataset, which is synthesized and optimized using Yosys, resulting in a gate-level netlist. This netlist is then converted into a graph structure during the graph generation phase. The next step involves node labelling, where each node, edge, and node label within the graph is identified and annotated. These labelled graphs are processed through a GCN followed by a Softmax activation function to extract features and classify the nodes. The processed graph data, including nodes, edges, and their labels, is then used for training a ML model. The trained ML model is subsequently employed to detect HT-infected nodes, providing detection results that distinguish between normal and HT-infected nodes. This method leverages the structural information of the graph and the learning capability of ML models to effectively identify and classify potential hardware Trojans in the dataset.

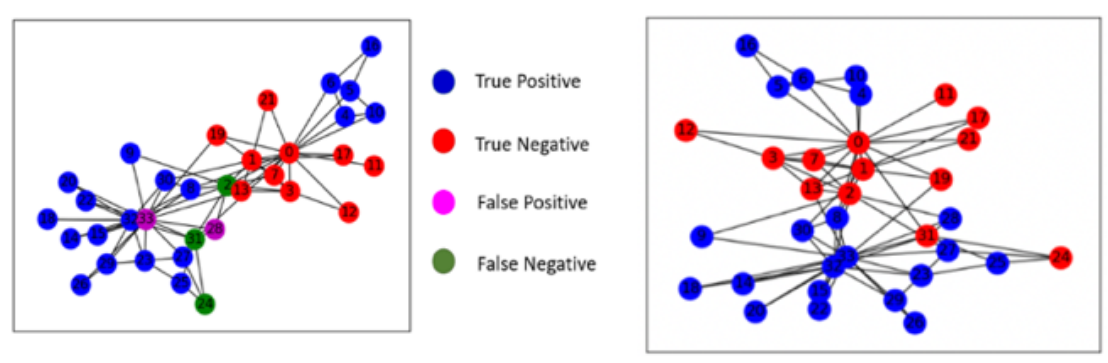

**Fig. 3**. Node classification in netlist

In the domain of HT detection within gate-level netlists, leveraging GNNs for node classification has emerged as a promising approach. Node-to-node classification involves analyzing the entire netlist graph to discern whether it exhibits characteristics indicative of a HT. Meanwhile, node classification focuses on identifying individual nodes within the netlist that may be compromised or suspicious. Further details are provided in Algorithm 3. Additionally, the model is integrated with 1st and 2nd NN approaches as discussed in Case II. Figure 3 illustrates the node classification performance of a GNN model, where nodes are classified into four categories: blue for true positives, red for true negatives, pink for false positives and green for false negatives. The graph highlights how the GNN model effectively distinguishes between Trojan infected and non-infected nodes within the netlist.

**Algorithm 3: GNN Model for Node Classification in Trojan Detection**

```
Input: Gate-level Netlist files (Verilog) of circuits (.v)
Output: Trojan Infected Node(s) or Trojan-Free Node(s)
def trojan_detection_node_classification (netlist_files):
    infected_nodes = [ ]  # To store infected nodes
    for each circuit.v in netlist_files do:
        g ← HW2GRAPH (.v)
        hg ← GRAPH2VEC (g)
        ŷ ← GCN (hg)
        for node in g.nodes:
            ŷ_node = GCN_node_classification (node, hg)

            if ŷ_node == 1:
                infected_nodes.append (node)

        if infected_nodes:
            post_process_infected_nodes (infected_nodes, circuit)
            return f "TROJAN Detected in nodes: {infected_nodes}"

    return "NON_TROJAN"

# Post-Processing Function
def post_process_infected_nodes (infected_nodes, circuit):

/*
   Maps infected graph nodes back to the gate-level netlist.
   Prints infected node locations.

   Args:
      infected_nodes (list): Nodes flagged as Trojan-infected.
      circuit (str/object): Source netlist for mapping.

   Returns:
      None
   */
    for node in infected_nodes:
        node_in_netlist = find_node_in_netlist (node, circuit)
            print (f "Infected node found in netlist: {node_in_netlist}")

NN1 = find_nearest_neighbour(node, level=1)
NN2 = find_nearest_neighbour(NN1, level=2)
f_i = extract_features(node, NN1, NN2)
y_node = GCN_node_classification(f_i)
```

## 5. Results and Discussions

The experiments are carried out on a Workstation with an Intel I Core I i7-4770 CPU, 16 GB of RAM, and Ubuntu 20.04.1 as the operating system.

*Case I: HT detection using machine learning mode*

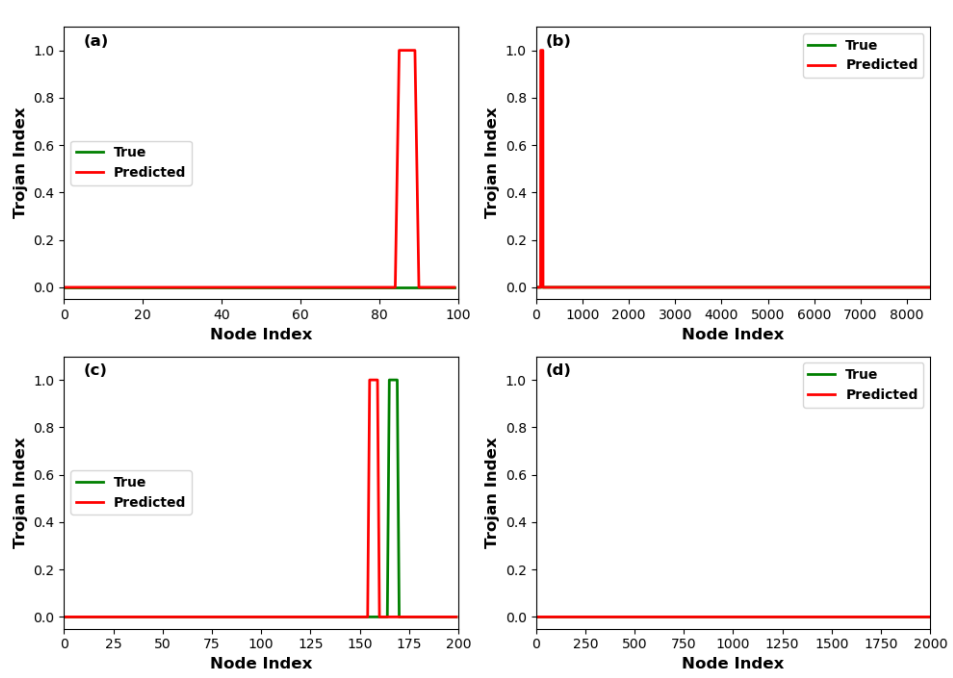

**Fig. 4.** HT detection result using decision tree algorithm for four random cases of test data sets.

The study compared the performance of HT detection with and without PCA across various input shapes, as shown in Table 1. For input shapes of 226, 808, and 1025, the dataset comprised 128, 593, and 711 samples with Trojans and 98, 215, and 314 samples without Trojans, respectively. The data is split with an 80%:20% between training and test data. Without PCA, the accuracy achieved was 35%, 41%, and 41% for the respective input shapes. In contrast, applying PCA resulted in significant improvements, yielding accuracies of 42%, 98.3%, and 97.54%. These findings indicate that PCA effectively enhances the model's performance, particularly for larger input shapes, by reducing dimensionality while preserving critical features essential for accurate Trojan detection. Four random sample's outputs are shown in Figure 4, where three cases have Trojan and one is without a Trojan. It can be seen that the model predicts the Trojan accurately in all the four cases. However, in one case (i.e., in Figure 4(c)), the node index is not predicted correctly.

Table 1

Classification performance comparison with and without PCA

| **Parameters** | **without PCA** | | | **with PCA** | | |
|---|---|---|---|---|---|---|
| Input shape | 226 | 808 | 1025 | 226 | 808 | 1025 |
| With Trojan | 128 | 593 | 711 | 128 | 593 | 711 |
| Without Trojan | 98 | 215 | 314 | 98 | 215 | 314 |
| Accuracy | 35% | 41% | 41% | 42% | 98.3% | 97.54% |

*Case II: GNN model using graph to graph classification*

In this section, a GNN model with graph-to-graph classification is presented. Firstly, the details of various nodes used in this study are explained. Figure 5 compares a network's (a) true structure with (b) a predicted structure, where red nodes are infected outputs and blue nodes are features. Green nodes in the predicted graph indicate false negatives, meaning the model missed these connections during prediction. Figure 5(c) isolates the infected output nodes, highlighting the nodes in the network that are predicted as compromised. This is done for the train dataset. Figure 6 provides a more detailed view of a graph-based HT detection for the test data set, highlighting three categories of nodes. Red nodes represent target nodes, blue nodes indicate infected nodes, green nodes indicate false negative nodes and yellow nodes denote the 1st NN Nodes. This visualization emphasizes the use of the NN algorithm to enhance Trojan detection by expanding the scope of analysis beyond just the infected nodes. The inclusion of 1st NN nodes suggests a broader capture of potentially compromised areas in the hardware, further improving the accuracy and coverage of the detection model. This model reinforces the value of the NN algorithm in augmenting graph-based detection models by identifying nodes closely connected to infected ones, increasing the likelihood of uncovering hidden or adjacent threats within the circuit.

The GNN model with graph-to-graph shows an accuracy of 62.8%, as shown in Table 2. Subsequently, 10 benchmark circuits have been considered for further data analyses. It is observed that the GNN model's average code coverage and maximum code coverage are 3.54% and 6.7% for these 10 benchmark circuits, respectively (refer Figure 7 and Table 2 for further details).

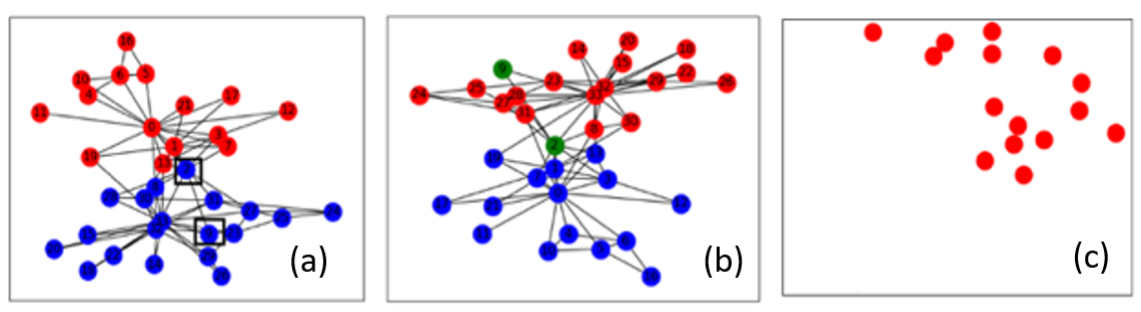

**Fig.5.** GNN model prediction for training data set (a) true graph, (b) predicted graph, and (c) infected nodes (target nodes (Red), infected nodes (blue), and false negative nodes (green))

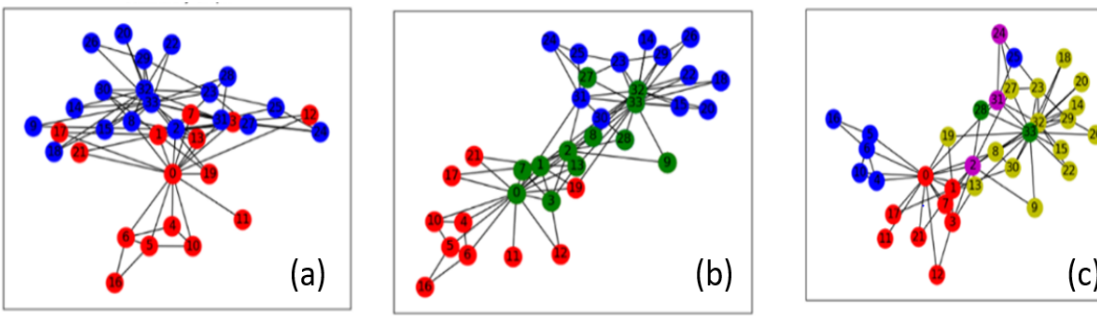

**Fig.6.** GNN model prediction for the testing data set including 1st NN nodes (a) true graph, (b) predicted graph, and (c) infected nodes (target nodes (Red), infected nodes (blue), False Negative nodes (green), and 1st NN nodes (yellow)).

Next, both the 1st and 2nd NN methods have been used to check the improvements in the accuracy of the basic GNN graph-to-graph model as shown in figure 7 graph -graph prediction model, (a) the 1st NN enhances prediction accuracy, closely matching true benchmark values and (b) the 2nd NN captures broader structural context, resulting in higher sensitivity across benchmarks. Details are presented in Table 2. Results shows that the 1st NN method improves the model accuracy from 62.8% to 73.2%. However, it has to be noted that the average code coverage and maximum code coverage are 5.0% and 7.8% for these 10 benchmark circuits, respectively (refer Figure 7 and Table 2 for further details). It leads to minimum time saving of ~92% while finding out the HTs inside the circuits. When comparing the model's performance with the 2nd NN approach, the GNN model showed an improved prediction accuracy of 97.7%. The average and maximum code coverage of GNN model with 2nd NN approach for 10 benchmark samples are 32.1% and 50.0%, respectively. The 2nd NN approach leads to minimum time saving of ~50% while finding out the HTs inside the circuits. These results highlight that while the GNN model shows promising results, the NN approaches, deliver superior performance in terms of accuracy with significant amount of time saving while identifying the exact locations of the HTs inside the circuits. Also, it can be noted that the average time for model to predict the HTs is very less (refer Table 2 for further details). Hence, the GNN graph-to-graph model along with the NN approach (either 1st or 2nd) can be used intelligently to exactly pin-point the HT locations in a large complex IC circuits.

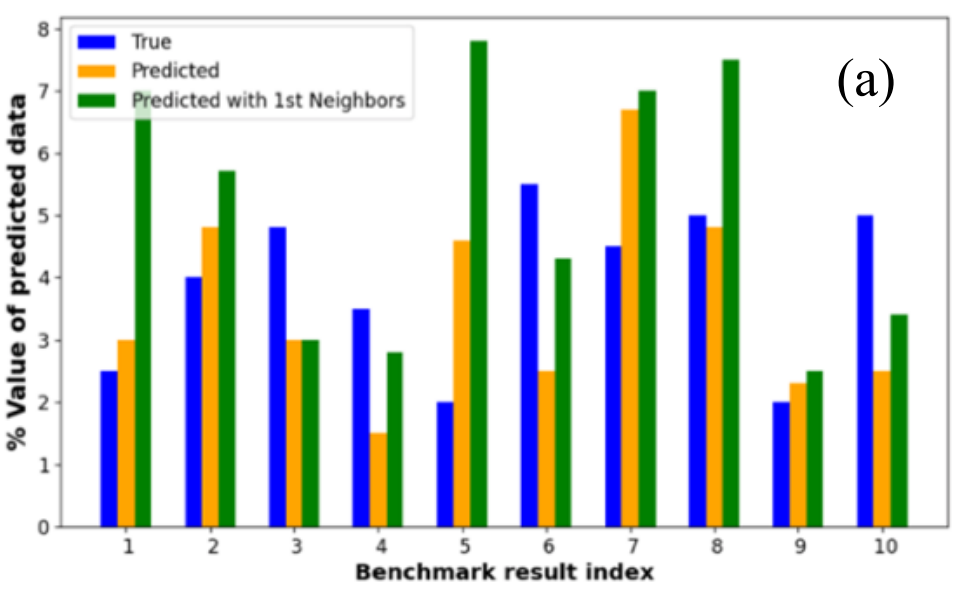

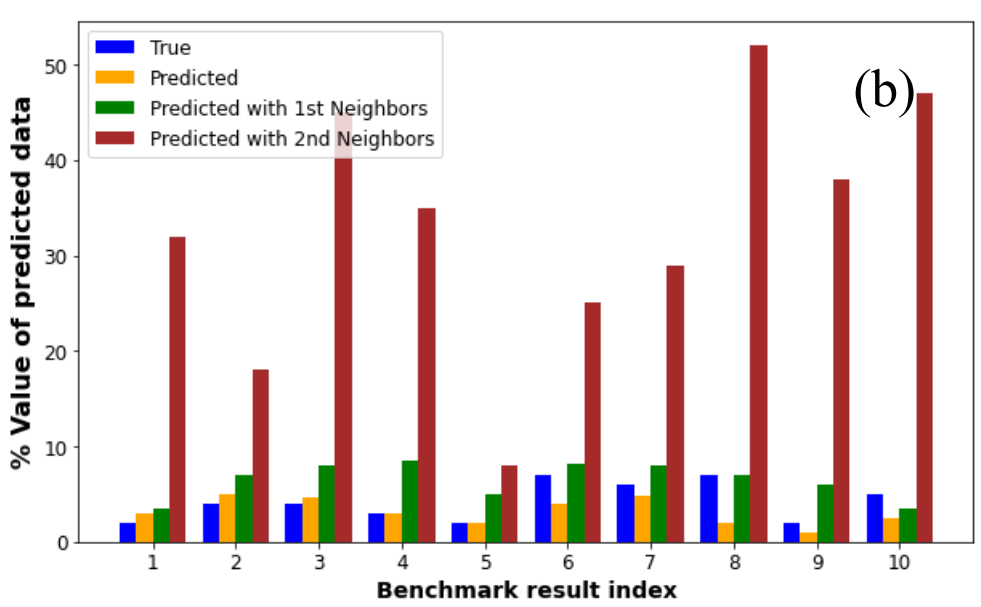

**Fig.7.** Graph- Graph prediction model (Case II), (a) Prediction with 1st NN and (b) Prediction with 2nd NN.

Table 2

Accuracy and code coverage for graph-graph (G-G) prediction model (Case II)

| | Accuracy | Average Code Coverage (%) | Maximum Code Coverage (%) | Average detection time(s) | Minimum time saved for HT detection (%) |
|---|---|---|---|---|---|
| GNN G-G | 62.8 | 3.5 | 6.7 | 0.805 | ~93% |
| GNN G-G with 1st NN | 73.2 | 5.0 | 7.8 | 0.430 | ~92% |
| GNN G-G with 2nd NN | 97.7 | 32.1 | 50.0 | 0.302 | ~50% |

Table 3

Performance of proposed model for graph-to-graph trojan detection (Case II).

| **Input Layer** | **Hidden layers** | **Classification** | **Optimizer** | **Learning Rate** | **No of epochs** | **Loss function** | **Output Layer** | **Test Accuracy (%)** | **Accuracy (%) with NN (known)** | | **Accuracy (%) with 1st NN (Unknown)** |
|---|---|---|---|---|---|---|---|---|---|---|---|
| | | | | | | | | | **1st NN** | **2nd NN** | |
| Adjacency matrix of DFG | 12 | Softmax | Adam | 0.001 | 200 | Cross entropy | Infected Nodes | 56.7 | 72.0 | 83.0 | 85.2 |
| Adjacency matrix of DFG | 16 | Softmax | Adam | 0.001 | 200 | Cross entropy | Infected Nodes | 62.8 | 73.2 | 91.7 | 92.7 |
| Adjacency matrix of DFG | 16 | Softmax | AdamW | 0.001 | 200 | Cross entropy | Infected Nodes | 59.3 | 66.8 | 84.9 | 83.8 |
| Adjacency matrix of DFG | 16 | Softmax | Adam | 0.001 | 250 | Cross entropy | Infected Nodes | 61.6 | 72.4 | 90.4 | 91.2 |
| Adjacency matrix of DFG | 16 | Softmax | Adamax | 0.001 | 250 | Cross entropy | Infected Nodes | 58.5 | 69.8 | 85.9 | 87.4 |

Details of ML parameters including input layer, hidden layers, classification, optimizer, learning rate, number of epochs, loss function, output layers are shown in Table 3 for GNN node-to-node classification approach. A large number of parametric studies are carried out and finally, model with 16 hidden layers, Softmax classification, Adam optimizer, learning rate of 0.001 with 200 epochs, and loss function of cross entropy is considered.

*Case III: GNN model using node-to-node classification*

Figure 8 represents the node-to-node model prediction. (a) Incorporating the $1^{st}$ NN improves prediction consistency with actual benchmark values and the $2^{nd}$ NN significantly amplifies predicted values, highlighting the effect of extended neighbourhood influence. In the evaluation of the GNN model using node-to-node classification, the model achieved a prediction accuracy of 79.8%. The GNN model's average code coverage is 3.9%, and its maximum code coverage is 5.4% for the 10 benchmark circuits considered, as shown in Figure 8 and Table 4. The 1st NN method leads to an improve in accuracy with 93.0%. Th average and maximum code coverage for the 1st NN model is 5.5% and 7.7%, respectively. It leads to minimum time saving of ~92% while finding out the HTs inside the circuits. When compared to the 2nd NN approach, which significantly outperformed with an accuracy of 97.7%. The average code coverage of 32.1% and a maximum code coverage of 50.0% is obtained for the 2nd NN approach. It leads to minimum time saving of ~50% while finding out the HTs inside the circuits. These results highlight that while the GNN model shows promising results, the nearest neighbour approaches, particularly the 2nd NN, deliver superior performance in terms of accuracy. Also, it can be noted that the average time for model to predict the HTs is very less (refer Table 4 for further details). Hence, the GNN node-to-node model along with the NN approach (either 1st or 2nd) can be used intelligently to exactly pin-point the HT locations in a large complex IC circuit. It has also to be noted that the GNN model with node-to-node (Case III) has better accuracy compared to GNN model with graph-to-graph (Case II) without integration of NN approach and also with the 1st NN approach.

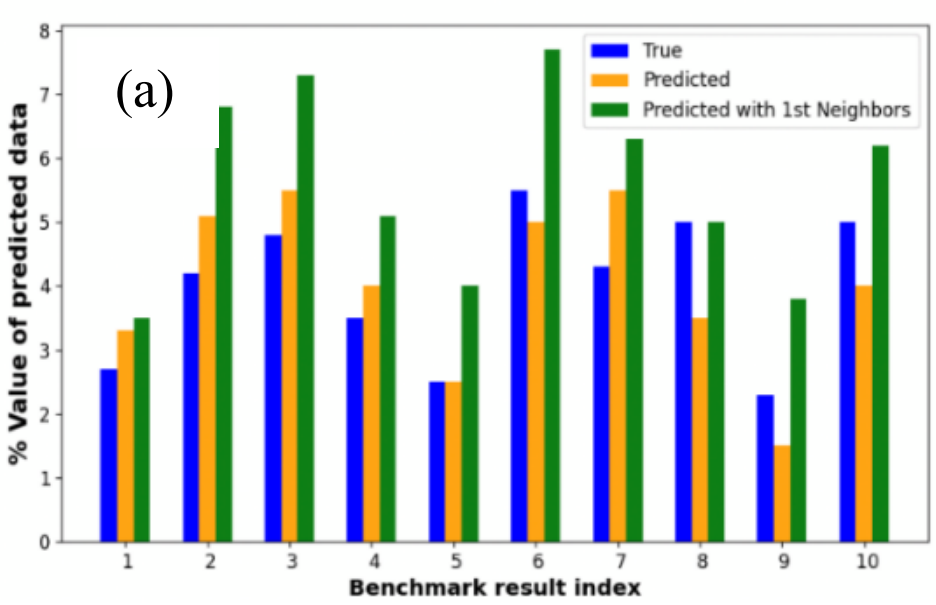

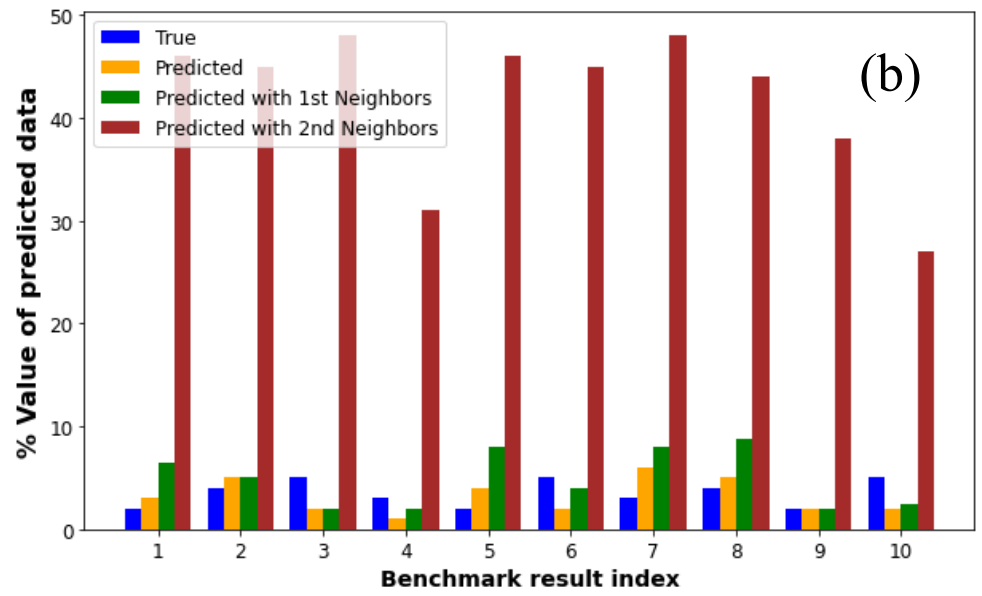

**Fig.8.** Node-Node prediction model (Case III),

(a) Prediction with $1^{st}$ NN and (b) Prediction with $2^{nd}$ NN.

Table 4

Accuracy and code coverage for node-node prediction model

| | Accuracy | Average Code Coverage (%) | Maximum Code Coverage (%) | Average detection time(s) | Minimum time saved for HT Detection (%) |
|---|---|---|---|---|---|
| GNN N-N | 79.8 | 3.9 | 5.4 | 0.840 | ~94% |
| GNN N-N with 1st NN | 93.0 | 5.5 | 7.7 | 0.510 | ~92% |
| GNN N-N with 2nd NN | 97.7 | 32.1 | 50.0 | 0.420 | ~50% |

Table 5 shows the parametric studies carried out for the GNN model with node-to-node classification. It can be seen that the adjacent matrix of DFG is used as an input layer along with various combinations of hidden layers, RELU and softmax classification. Different optimizers, i.e., Adam and AdamW, have been studied. The performance of the basic GNN model with node-to-node, with the $1^{st}$ NN (with known and unknown circuits), and with the $2^{nd}$ NN with known circuits has been evaluated. It can be seen that the NN along with a basic GNN model with node-to-node classification, helps to improve the accuracy, surpassing 90% in most cases. The highest accuracy achieved was 98.5% with 12 hidden layers, Softmax output, Adam optimizer, learning rate of 0.001, 250 epochs, and Categorical cross-entropy loss function. The Adam optimizer consistently outperformed AdamW, demonstrating its effectiveness for this task.

The time complexity has significant impact on the overall efficiency and effectiveness of the detection process. The GNN model and the $1^{st}$ NN network is responsible for processing the graph structure and extracting relevant features from the gate-level netlist. Its runtime complexity is O(N log N), i.e., the time it takes to process each node in the graph increases logarithmically with the number of nodes (N). Memory usage is low, making this network efficient for large-scale netlists. The $2^{nd}$ NN network uses the features extracted by the first NN to make predictions about potential Trojans in the netlist. Its runtime complexity is $O(N^2)$, indicating that the time it takes to process each node increases quadratically with the number of nodes (N). This network requires comparatively more memory to operate efficiently.

Table 5

Performance of proposed model for node-to-node trojan detection (Case III).

| **Input Layer** | **Hidden layers** | **Classification** | **Optimizer** | **Learning Rate** | **No of epochs** | **Loss function** | **Output Layer** | **Test Accuracy (%)** | **Accuracy (%) with NN (known)** | | **Accuracy (%) with $1^{st}$ NN (Unknown)** |
|---|---|---|---|---|---|---|---|---|---|---|---|
| | | | | | | | | | **$1^{st}$ NN** | **$2^{nd}$ NN** | |
| Adjacency matrix of DFG | 10 | Softmax | Adam | 0.001 | 250 | Categorical Cross entropy | Node Labels | 73.3 | 83 | 86.2 | 89 |
| Adjacency matrix of DFG | 12 | Softmax | Adam | 0.001 | 250 | Categorical Cross entropy | Node Labels | 79.8 | 93 | 97.7 | 98.5 |
| Adjacency matrix of DFG | 12 | Softmax | AdamW | 0.001 | 250 | Categorical Cross entropy | Node Labels | 74.6 | 91.8 | 96.3 | 94.3 |

The comparison of various state-of-the-art hardware Trojan detection models are also presented in Table 6. It highlights distinct trade-offs in detection accuracy, localization capability, and computational efficiency. Traditional models such as GNN4TJ [18] and DFF [20] demonstrate strong detection performance but lack localization abilities, limiting their effectiveness in pinpointing compromised nodes. Methods like FastText [22] and Decision Tree [23] incorporate localization but may suffer from lower recall or structural limitations. Advanced graph-based techniques, such as GCN [24] and node-wise detection models, offer improved node extraction and edge-level insights, enhancing Trojan identification. The proposed model achieves a strong balance, combining high recall with precise localization while maintaining efficient node and edge extraction. This comprehensive comparison underscores the evolving landscape of Trojan detection, where graph learning-based methods show significant promise in improving both detection accuracy and interpretability.

The Proposed Model outperforms existing methods in terms of precision 98.5%, while maintaining the highest recall 99.1% and strong F1-score 96.7%. The use of 2nd NN method contributes to improved detection performance. Also, it is to be noted that the largest number of data samples are used in the present work.

Table 6

Comparison with other state-of-art models.

| Method | Dataset used | No. of data samples | Detection | Localization | Precision | Recall (TPR) | F1-Score | Node Extraction | Edges Extraction | Nearest neighbour |
|---|---|---|---|---|---|---|---|---|---|---|
| GNN4TJ [18] | Trust-HUB | 100 | ✓ | ✗ | 92.3% | 96.6% | 94.0% | ✓ | ✓ | ✗ |
| GNN [19] | Trust-HUB + Custom HDL | 422 | ✓ | ✗ | 96.6% | 100.0% | 98.4% | ✓ | ✓ | ✗ |
| DFF [20] | Trust-HUB + Opencores+ Custom HDL | 98 | ✓ | ✗ | 97.8% | 93.8% | 95.5% | ✓ | ✓ | ✗ |
| FV [21] | Trust-HUB | 28 | ✓ | ✗ | - | - | - | ✗ | ✗ | ✗ |
| Fast-Text [22] | | | ✓ | ✓ | 97.9% | 97.8% | 97.5% | ✓ | ✓ | ✗ |
| Decision-Tree [23] | Trust-HUB | 52 | ✓ | ✓ | 100.0% | 78.8% | 88.1% | ✓ | ✓ | ✗ |
| GNN [24] | Trust-HUB | 132 | ✓ | ✓ | 94.5% | 99.1% | 96.7% | ✓ | ✓ | ✗ |
| Node-wise HT detection [25] | Trust-HUB | 24 | ✓ | ✓ | 95.0% | 98.0% | 96.4% | ✓ | ✓ | ✓ |
| **Proposed Model** | Trust-HUB+ web3 Hardware + Custom HDL | 1025 | ✓ | ✓ | 98.5% | 99.1% | 96.7% | ✓ | ✓ | ✓ |

## 6. Conclusions

In this research, a machine learning-based methodology to tackle the challenge of detecting Hardware Trojans (HTs) in gate-level netlists within the multi-entity design process of integrated circuits is proposed. By leveraging path retrace algorithms, the approach successfully identifies malicious nets introduced by adversaries, pinpointing their locations within the circuit. The effectiveness of the proposed methods arevalidation across three distinct cases, each employing different machine learning models. Case I showed that integrating PCA with decision tree algorithms improves the detection accuracy for node-to-node comparisons. Case II and Case III employed GNNs for graph-to-graph and node-to-node classifications, respectively, to detect HTs. The NN method, when combined with GNN-based models, further enhanced the detection performance. Notably, the 2nd NN method achieved the highest accuracy, outperforming GNN models with 97.7% accuracy in both graph-to-graph and node-to-node classifications. The proposed technique of integration of NN with base GNN models may further be explored in other domain as well for improving the machine learning models.